\documentclass[conference]{IEEEtran}

\usepackage{cite}
\usepackage{amsmath,amssymb,amsfonts}
\usepackage{algorithmic}
\usepackage{graphicx}
\usepackage{textcomp}
\usepackage{subcaption}
\usepackage{algorithm}
\usepackage{xcolor}
\def\BibTeX{{\rm B\kern-.05em{\sc i\kern-.025em b}\kern-.08em
    T\kern-.1667em\lower.7ex\hbox{E}\kern-.125emX}}
\begin{document}

\title{GAMORA: A Gesture Articulated Meta Operative Robotic Arm for Hazardous Material Handling in Containment-Level Environments\\}
\author{\IEEEauthorblockN{ Farha Abdul Wasay}
\IEEEauthorblockA{\textit{Department of Computer Science \& AI} \\
\textit{Muffakham Jah College of Engineering and Technology}\\
Hyderabad, India \\
farhawasay@gmail.com}
\and
\IEEEauthorblockN{Mohammed Abdul Rahman}
\IEEEauthorblockA{\textit{Department of Computer Science \& AI} \\
\textit{Muffakham Jah College of Engineering and Technology}\\
Hyderabad, India \\
writetoabdulrahman11@gmail.com}
\and
\IEEEauthorblockN{Hania Ghouse}
\IEEEauthorblockA{\textit{Department of  Computer Science \& AI} \\
\textit{Muffakham Jah College of Engineering and Technology}\\
Hyderabad, India\\
haniaghouse704@gmail.com}
}

\maketitle

\begin{abstract}
The rapid advancements in robotics and virtual reality (VR) technologies have paved the way for safer and more efficient laboratory environments, particularly in high-risk settings such as virology labs. The increasing complexity and bio-hazard risks in virology laboratories demand advanced solutions that minimize direct human exposure while maintaining high precision. To address this, we propose GAMORA (Gesture Articulated Meta Operative Robotic Arm), a novel VR-guided robotic system that enables remote execution of high-risk tasks via natural hand gestures.  Unlike existing teleoperation or scripted automation approaches, GAMORA uniquely integrates the Oculus Quest 2 VR interface, NVIDIA Jetson Nano, and Robot Operating System (ROS) to support immersive real-time control, digital twin simulation, and optimized robotic arm articulation using inverse kinematics. The system enables immersive virtual training and simulation for researchers, while a VR-controlled robotic arm performs precise, high-risk tasks, significantly reducing direct human exposure to infectious agents. Advanced inverse kinematics are implemented to ensure accurate robotic arm manipulation, critical for handling delicate samples. Through a detailed methodological framework, including 3D environment development, data processing, and rigorous testing and optimization, the system demonstrates its potential to transform virology lab operations by enhancing safety, efficiency, and precision. A fully simulated and physical 3D-printed arm was developed to evaluate the system. GAMORA achieved a mean 2.2 mm positional discrepancy (improved from 4 mm), and pipetting accuracy within 0.2 mL of the target volume, and repeatability of ±1.2 mm across 50 consecutive trials—demonstrating competitive performance in high-risk manipulation tasks. The proposed system not only enhances biosafety by decoupling the operator from hazardous zones but also offers a scalable framework for immersive training and robotic control in biomedical research settings.
\end{abstract}


\section{Introduction}
Virology laboratories play a critical role in infectious disease research, diagnostics, and response \cite{in1}\cite{in2}. However, these high-containment environments expose researchers to significant occupational risks, including accidental contamination, direct contact with hazardous biological agents, and repetitive strain injuries due to manual procedures \cite{in3}\cite{in4}. Despite strict biosafety protocols, reports indicate that nearly 70\% of infectious lab accidents stem from human error, such as pipetting, culturing, and sample transfer mishaps\cite{in5} \cite{in6}. The integration of robotic systems into biomedical laboratories has emerged as a promising solution to mitigate these risks \cite{in7}. Robotic arms can automate precision-driven processes, thereby enhancing reproducibility and reducing human exposure to infectious materials \cite{in8}. Nevertheless, their adoption remains limited in clinical and research settings, with only an estimated 5\% of hospitals globally deploying robotic solutions \cite{in9}. A key barrier is the lack of intuitive human-robot interaction (HRI) interfaces and cost-effective platforms that allow seamless integration into existing workflows\cite{in10}. In parallel, immersive technologies such as augmented reality (AR) and virtual reality (VR) are reshaping the landscape of medical training and procedural simulation\cite{in11}. VR-based platforms have been shown to improve user engagement, reduce cognitive load, and offer safe, high-fidelity environments for skill acquisition \cite{in12}. The increasing availability of low-cost, standalone VR headsets—such as the Oculus Quest 2—has further democratized access to immersive systems \cite{in13}. These devices offer advanced hand-tracking, high-resolution displays, and untethered interaction capabilities, making them suitable candidates for gesture-based control of robotic systems \cite{in14}.  VR systems, when used in conjunction with laboratory robotics and AR, allow researchers to practice in virtual environments that replicate real-world complexities without the associated risk \cite{in15}. The integration of Robot Operating System (ROS) and Jetson Nano in 5-Degree Of Freedom (DOF) robotic arms has been effectively demonstrated in various research applications, particularly in laboratory settings \cite{in16}.  However, existing VR-robotic interfaces are often constrained to training-only environments or limited to industrial use cases \cite{in17}.

\section{Literature Review}
Advancements in virtual reality (VR), mixed reality (MR), and robotics have significantly contributed to automation, teleoperation, and training in high-risk biomedical environments \cite{lr1} \cite{lr2}. Several studies have explored the integration of robotic systems to enhance biosafety and reduce manual handling of infectious samples \cite{lr18} \cite{lr19} \cite{lr20}. Angers et al.~\cite{lr3} proposed an automated pipeline for sample preparation and testing, minimizing direct exposure for laboratory personnel. However, their closed-loop design lacked flexibility, as each system required hardware-specific adaptation for different pathogens. Moreover, manual sample insertion was still necessary, limiting full autonomy. To support real-time robotic simulation and interaction, Babaians et al.~\cite{lr4} introduced a Unity3D \cite{lr11} based platform with ANet middleware, employing ZeroMQ and Google Protobuf for efficient communication. Unity–ROS integration has proven highly effective for real-time robotic simulation, with Platt et al. \cite{lr12} and Kuts et al. \cite{lr13} highlighting its low latency and high precision in digital twin frameworks .While this enabled the generation of realistic robotic sensor data for training AI models, it introduced complexity in software dependencies and required continual updates to remain stable. Chinnasamy et al.~\cite{lr5} and Dey et al. \cite{lr14} implemented a ROS–Gazebo digital twin and identified significant synchronization lags between virtual and physical models, resulting in reduced motion repeatability—highlighting the need for more robust closed-loop architectures. Lundeen et al.~\cite{lr6} reviewed ROS–Unity and VR integrations, demonstrating that stability and timing drift continue to limit reliability in immersive teleoperation systems. In parallel, immersive control interfaces have gained attention for improving human-robot interaction \cite{lr15} \cite{lr16}. Rosen et al.~\cite{lr7} demonstrated that MR-based head-mounted displays (HMDs) can reduce task completion time compared to conventional 2D interfaces. However, positional drift due to SLAM inaccuracies was found to degrade control reliability during prolonged use. Tsai et al.~\cite{lr8} applied VR simulations to virological testing, enhancing learning engagement but noted that developing high-fidelity 3D assets was labor-intensive and resource-heavy, posing challenges for scalability. Despite substantial progress in robotic motion planning, Chan et al.~\cite{lr9} identified persistent challenges in achieving precise trajectory control, especially in environments with curved paths or tight spatial constraints. Santos et al.~\cite{lr10} highlighted the inefficiency of manual robot programming via Teach Pendants \cite{lr17}, further reinforcing the need for automated, adaptive interfaces. Collectively, these studies demonstrate the potential of VR and robotics to transform laboratory practices, but also reveal several limitations—namely, the lack of real-time adaptive learning, dependence on manual intervention, and limited flexibility in constrained environments. Existing systems often separate training from execution, rely on scripted control, or incur high development costs for immersive content. Hence to overcome these limitations, we propose GAMORA: a Gesture-Articulated Robotic Arm system that integrates VR-based control, ROS infrastructure, and reinforcement learning to enable real-time, user-guided automation in biosafety lab environments.

\section{Proposed System}
\begin{figure}[htbp]
\centerline{\includegraphics[width=\linewidth, height=8cm, keepaspectratio]{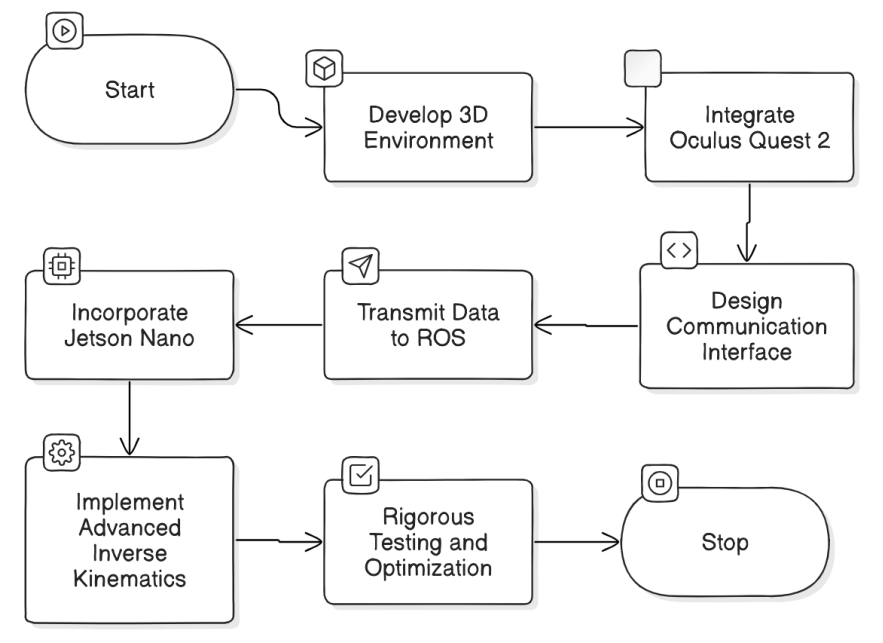}}
\caption{System Workflow}
\label{fig1}
\end{figure}

To bridge the gap in VR-enabled robotic systems for biosafety lab environments, we present a novel architecture shown in Fig.\ref{fig1}, designed to enhance safety, precision, and training in virology laboratories. The workflow begins with the development of a 3D virtual lab environment, enabling immersive simulations. The Oculus Quest 2 is integrated to provide gesture-based interaction and control, allowing users to operate the system in a risk-free, virtual setting. Jetson Nano serves as the onboard processor, managing sensor data and control commands. The Robot Operating System (ROS) facilitates seamless communication between the VR interface and the robotic arm. Data from user actions is transmitted in real-time to execute high-precision tasks using inverse kinematics. This modular pipeline ensures accurate, repeatable manipulation of infectious samples while minimizing direct human exposure and improving training scalability.

\subsection{Hardware SetUp}
\begin{figure}[htbp]
\centerline{\includegraphics[width=\linewidth, keepaspectratio]{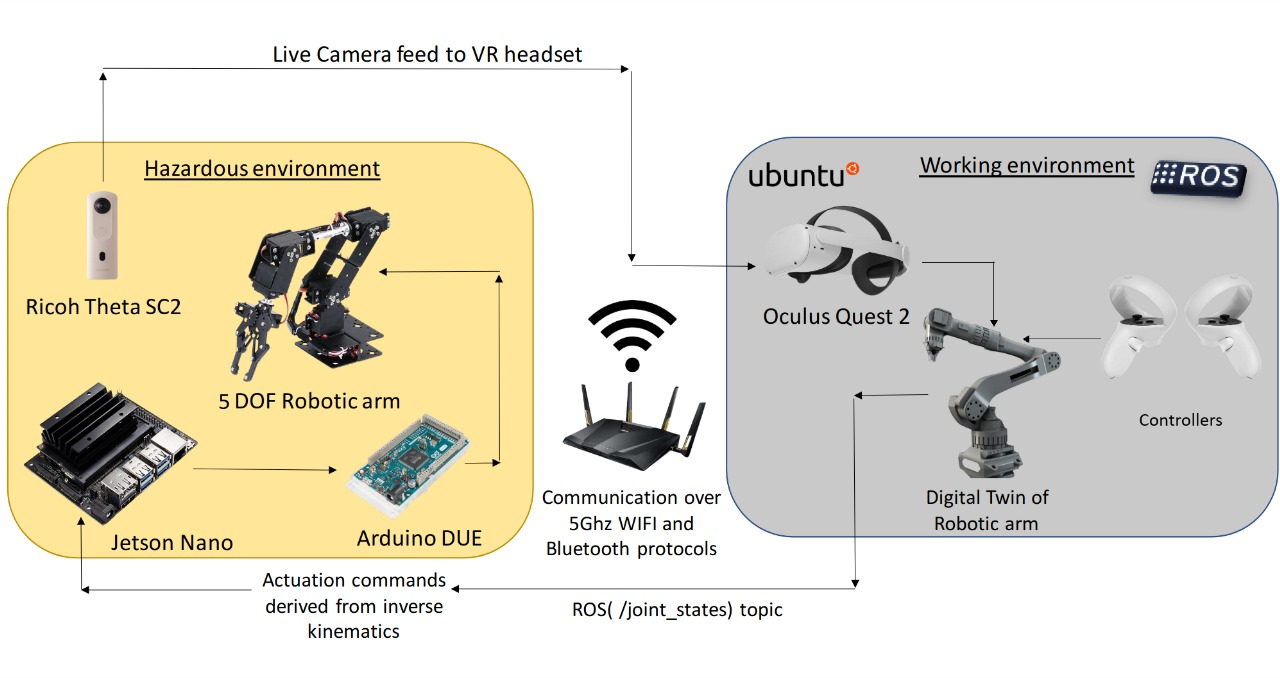}}
\caption{Hardware Setup}
\label{fig2}
\end{figure}

The proposed hardware setup, shown in Fig.\ref{fig2}, is designed to safely link the hazardous lab environment with the operator's remote interface. Within the hazardous zone, a 5-DOF robotic arm performs critical handling of infectious materials, guided by commands processed on a Jetson Nano and actuated via an Arduino DUE. A Ricoh Theta SC2 camera provides live video feedback to the operator through the Oculus Quest 2 VR headset, enabling immersive monitoring without physical exposure. In the operator’s environment, a digital twin of the robotic arm runs in ROS on Ubuntu, synchronized with the physical arm via the /joint\_states topic. Communication between environments is achieved through a 5GHz WiFi and Bluetooth connection. This integrated setup supports real-time control and enhances biosafety, while also ensuring system scalability and long-term reliability in virology lab automation.
\section{Methodology}

\subsection{Creation of the Virtual Environment}
\begin{figure}[htbp]
\centerline{\includegraphics[width=\linewidth, height=6cm, keepaspectratio]{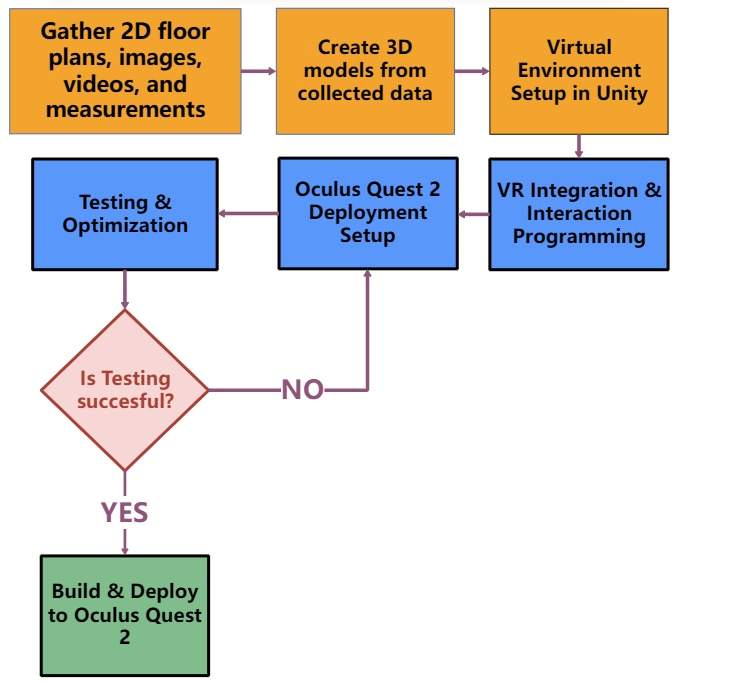}}
\caption{Process flow for creating virtual workspace}
\label{fig3}
\end{figure}

\begin{figure}[htbp]
\centerline{\includegraphics[width=\linewidth, height=5cm, keepaspectratio]{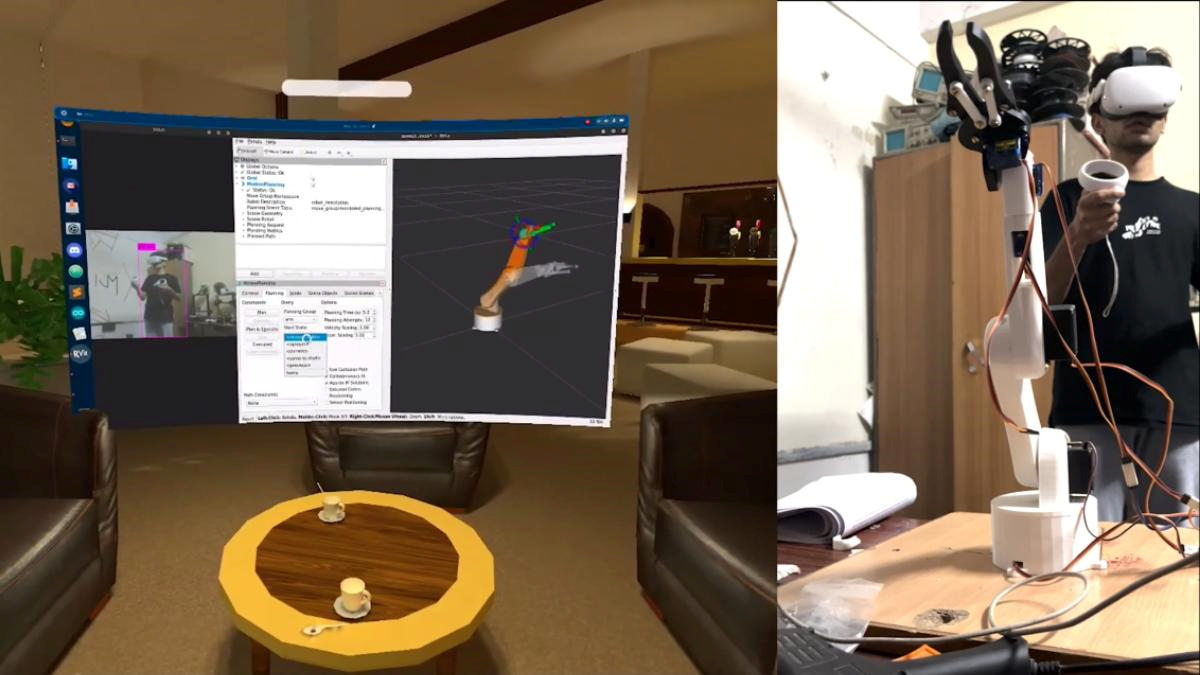}}
\caption{Virtual Workspace}
\label{fig4}
\end{figure}

The virtual environment for GAMORA was developed following a structured pipeline, as illustrated in Fig.~\ref{fig3}, aimed at achieving a high-fidelity, interactive simulation space suitable for virology lab training and teleoperation.
We began by collecting detailed spatial references—2D floor plans, images, and physical measurements—from the target lab environment. These inputs informed the creation of accurate 3D models representing essential workspace elements. Once generated, the 3D assets were imported into Unity 3D, ensuring compatibility with the rendering engine and physics system. Objects were then positioned precisely using Unity’s coordinate system to match their real-world spatial arrangement. The virtual layout was constructed with a modular approach—grouping elements based on functional zones or task relevance to ensure clarity and ease of navigation. With the physical structure replicated, we proceeded to implement interaction features. Custom scripts in C\# enabled user navigation and hand-based manipulation of virtual components. The Oculus Integration SDK for Unity was employed to support full VR functionality, including spatial tracking, gesture recognition, and controller input. The Oculus Quest 2 was selected as the target platform due to its high-resolution display, untethered operation, and robust hand-tracking capabilities. After initial deployment, the environment underwent iterative testing and optimization to ensure system stability, performance consistency, and accurate interaction mapping. The final result, shown in Fig.~\ref{fig4}, is a fully immersive and interactive virtual lab environment that supports real-time control and training simulations.

\subsection{Configuration of the Robotic Arm}
\begin{figure}[htbp]
\centering
\begin{subfigure}[b]{0.48\linewidth}
    \includegraphics[width=\linewidth, height=3cm]{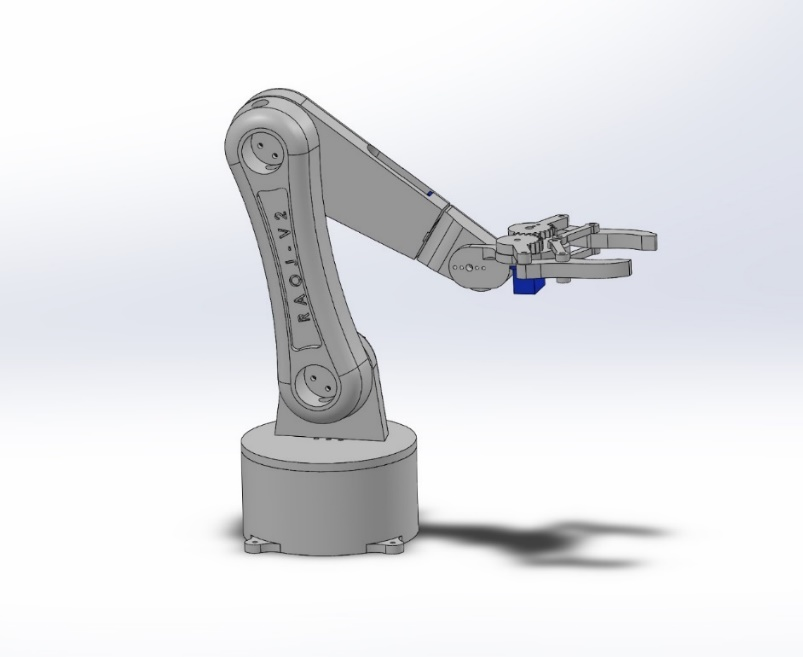}
    \caption{}
    \label{fig5a}
\end{subfigure}
\hfill
\begin{subfigure}[b]{0.48\linewidth}
    \includegraphics[width=\linewidth, height=3cm]{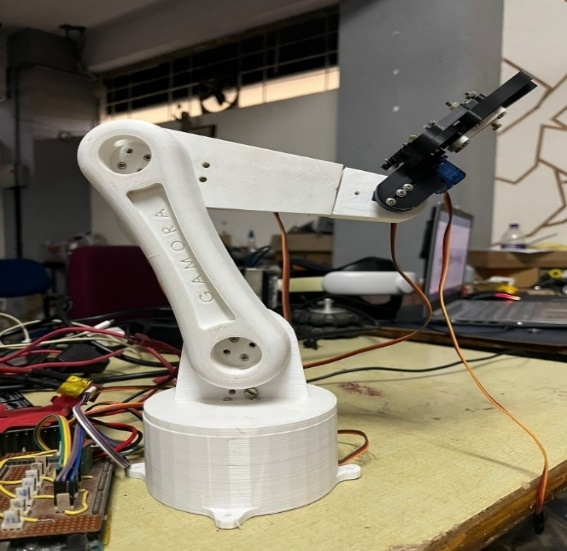}
    \caption{}
    \label{fig5b}
\end{subfigure}
\caption{GAMORA’s robotic arm: (a) digital design and (b) 3D-printed of robotic arm}
\label{fig5}
\end{figure}

The design of the GAMORA robotic arm followed a comprehensive digital-to-physical workflow, as shown in Fig.\ref{fig5}. The arm was first modeled using SolidWorks to define its kinematic structure, joint constraints, and end-effector geometry , as shown in Fig.\ref{fig5a}. The components were then fabricated using 3D printing, shown in Fig.~\ref{fig5b}, ensuring accurate translation of the CAD model into a functional prototype.
The arm utilizes high-torque RDS5180 servo motors (80 kg·cm) to support precision manipulation and payload handling. A Jetson Nano 4GB serves as the primary control unit, enabling real-time computation, while a DC-DC converter provides regulated power to all electronic subsystems. ROS was employed as the middleware framework to manage communication, control, and data flow. The robotic structure was defined using a Unified Robot Description Format (URDF), specifying link dimensions, joint types, and spatial relationships. ROS nodes were configured for motor control, trajectory planning, and sensor integration. RViz was used for real-time 3D visualization and debugging of motion planning and robot states, allowing for accurate alignment between the simulated and physical arm. This setup ensured modularity, precision, and repeatability—critical for remote operation in biosafety-sensitive environments.
\subsection{Motion Planning and Initial Calibration}
\begin{figure}[htbp]
\centerline{\includegraphics[width=\linewidth, keepaspectratio]{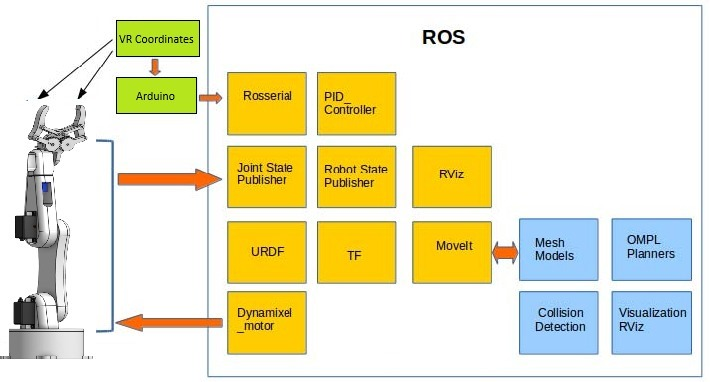}}
\caption{Hardware and software architecture with ROS packages}
\label{fig7}
\end{figure}

\begin{figure}[htbp]
\centerline{\includegraphics[width=\linewidth, keepaspectratio]{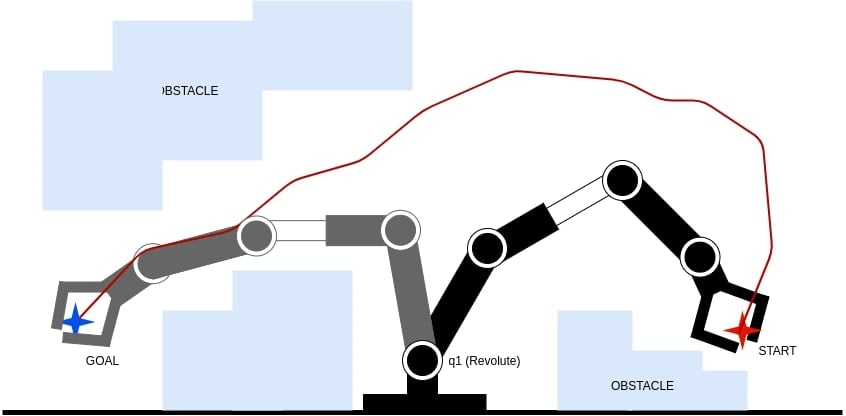}}
\caption{Path planning in robotic arm using Inverse Kinematics}
\label{fig8}
\end{figure}

Motion planning for GAMORA was implemented using the MoveIt! framework within ROS, as shown in Fig.~\ref{fig7}. MoveIt! enabled the generation of feasible, collision-free trajectories based on the robot's URDF-defined kinematic model. Planning algorithms such as RRT (Rapidly-exploring Random Tree) and PRM (Probabilistic Roadmap Method) were used to explore the configuration space and determine optimal joint paths for reaching target poses.
Inverse kinematics (IK) played a key role in this process, allowing the computation of joint angles required to position the end effector at a specific Cartesian coordinate. Fig.~\ref{fig8} illustrates the IK-based trajectory planning in an obstacle-filled workspace. MoveIt! utilized solvers like KDL to iteratively solve for joint angles while minimizing positional error and avoiding collisions.
The following equations govern the computation of joint angles $\theta_1$ and $\theta_2$ for a planar 2-DOF manipulator:
\begin{equation}
\cos \theta_2 = \frac{(x^2 + y^2 - L_1^2 - L_2^2)}{2 L_1 L_2} \tag{1}
\end{equation}
\begin{equation}
\sin \theta_2 = \mp \sqrt{1 - \cos^2 \theta_2} \tag{2}
\end{equation}
\begin{equation}
\theta_2 = \text{atan2}(\sin \theta_2, \cos \theta_2) \tag{3}
\end{equation}
\begin{equation}
\theta_1 = \text{atan2}(y, x) - \text{atan2}(L_2 \sin \theta_2,\ L_1 + L_2 \cos \theta_2) \tag{4}
\end{equation}
\noindent where $x$, $y$ are the target coordinates of the end effector, and $L_1$, $L_2$ are the lengths of the arm segments. These equations allow the robot to compute joint angles dynamically based on hand gestures transmitted via the Oculus Quest 2 VR interface. The computed joint trajectories are then executed in real-time, ensuring safe and precise manipulation within constrained environments.
\subsection{Object Detection Integration}
Object detection was integrated into the GAMORA system using YOLOv8 with default pretrained weights. This allowed for the identification of objects and features within the virtual or physical workspace. The detected objects helped enhance the spatial awareness of the robotic arm and the user, particularly in planning interactions and verifying the correct placement or retrieval of items. This AI component complemented the VR-based control system by providing contextual understanding of the environment without requiring manual annotation or training. Finally, the robotic arm's motion planning was tested and visualized in a simulated environment using tools such as RViz. By integrating the URDF model into RViz, we were able to visualize the arm’s kinematic structure and monitor its joint states and trajectory execution throughout the planning process. RViz enabled real-time observation of the paths generated by MoveIt!, helping evaluate how the robotic arm would interact with its environment. This allowed for effective validation of inverse kinematics, collision avoidance, and trajectory planning in a dynamic and realistic virtual workspace.
\section{Experimental Setup and System Testing}
\subsection{Calibration Experiments}
Calibration experiments for GAMORA evaluated visualization modes, control methods, and end-effector tools to optimize user interaction and task precision. Three visualization modes were tested. The basic view (no camera) offered minimal feedback, while a webcam provided clarity but lacked depth perception. A depth-sensing camera proved most effective, offering virtual zoom and enhanced spatial awareness—significantly improving operator immersion and accuracy in manipulation tasks. For control input, traditional mouse and keyboard lacked spatial intuitiveness. Hand-tracking via Oculus Quest 2 was tested but found imprecise for fine tasks. The Oculus Controllers delivered the best performance, offering real-time responsiveness and precise control of the arm’s end effector, making them the preferred interface for GAMORA. Two end-effector tools were evaluated: a suction cup and a gripper. While the suction cup handled flat, lightweight items well, it required pneumatic support and struggled with heavier loads. The gripper was better suited for small objects like vials but lacked force feedback, limiting its use in delicate operations. These results highlight the need for further refinement of adaptive end-effector systems for diverse lab tasks.

\subsection{Iterative Testing, Optimization, and System Refinement}
To validate GAMORA’s real-world performance, hardware-in-the-loop (HIL) testing was conducted by integrating the physical robotic arm with its ROS-based control software. This allowed the system to be evaluated under simulated deployment conditions without risking hardware damage. Sensor feedback, motor commands, and trajectory data were processed in real time to analyze the arm’s behavior and controller precision. Key performance metrics included joint angle accuracy, trajectory tracking, and control latency. Discrepancies between simulated and physical outcomes were used to iteratively refine both the kinematic model and control algorithms. Inverse kinematics solvers were tuned to reduce positional error, and control loop frequencies were optimized to enhance responsiveness. Sensor sampling rates and motion planning parameters were adjusted for improved real-time actuation. System debugging involved monitoring CPU/GPU utilization, ROS node latencies, and network performance. RViz was used extensively for visualizing joint states, sensor feedback, and system errors. Real-time updates to the URDF model and diagnostic feedback enabled continual calibration, ensuring stable and precise operation in safety-critical environments.
\subsection{Real-Time Feedback Integration and Final System Implementation}
Real-time feedback in GAMORA was achieved by continuously monitoring sensor inputs—including joint encoders, force-torque sensors, and visual data from onboard cameras—and dynamically adjusting the robotic arm’s behavior through ROS. This allowed the system to respond adaptively to environmental changes and task requirements, improving precision and safety. The final system integration brought together all subsystems—hardware, control software, feedback mechanisms, and planning algorithms—into a unified and functional architecture. Actuators, sensors, and controllers were fine-tuned to operate seamlessly within the ROS ecosystem. Motion planning was executed using MoveIt!, while inverse kinematics solvers ensured accurate end-effector positioning. Full-system testing validated the synchronization between hardware and software, confirmed the system’s responsiveness, and resolved latency or feedback issues using ROS diagnostic tools. The integrated platform was assessed across multiple tasks, including object handling and assembly, demonstrating high stability, accuracy, and fault tolerance. These results confirmed GAMORA’s readiness for deployment in both laboratory automation and industrial settings.
\section{Results and Discussion}
Pilot trials were conducted to assess the operational performance of the GAMORA system in critical virology laboratory workflows. Tasks such as specimen manipulation, pipetting, and sample transfer—typically requiring high precision and reproducibility—were used to evaluate system effectiveness. The robotic arm, controlled remotely via Oculus Quest 2 VR and equipped with a gripper end-effector, successfully handled delicate glass vials and executed precise fluid transfer. The system demonstrated consistent control stability and task accuracy across multiple stages, confirming its potential to reduce manual intervention and support safe, repeatable procedures in high-containment laboratory settings. The experiments were performed in several stages, each concentrating on various laboratory tasks:
\subsection{Specimen Handling and Placing}
The GAMORA system was evaluated for its ability to handle and accurately place standard glass viral culture vials using a gripper-based end-effector. The system could reliably manipulate vials up to 20 grams, meeting the requirements for typical specimen containers used in virology labs. Performance metrics were as follows:
\begin{itemize}
\item \textbf{Positional Accuracy:} Improved from 4.0 mm to 2.2 mm after calibration.
\item \textbf{Angular Accuracy:} Reduced misalignment from 8.5° to 2.5° during vial insertion.
\item \textbf{Repeatability:} Achieved ±2.8 mm deviation across 20 placement cycles.
\end{itemize}
These results indicate that GAMORA is capable of precise and repeatable specimen handling in virology lab conditions.
\subsection{Pipetting and Liquid Handling}
\begin{figure}[htbp]
\centerline{\includegraphics[width=\linewidth, keepaspectratio]{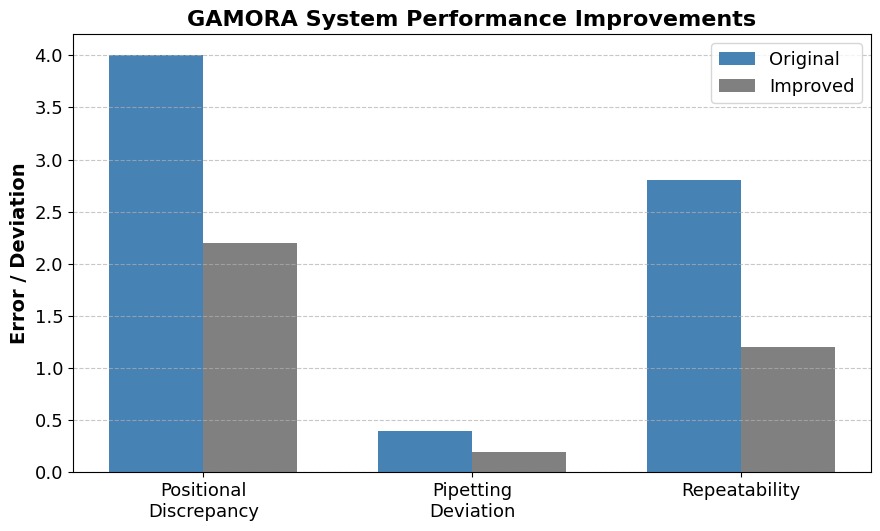}}
\caption{Comparative performance metrics of the original and improved GAMORA system.}
\label{fig9a}
\end{figure}

GAMORA was evaluated on VR-guided liquid handling tasks, involving pipetting between containers with varying liquid viscosities and geometries. The system achieved an average pipetting deviation of 0.2mL from a 1mL target, with a repeatability of ±0.1mL and a 95\% success rate. Initial pipette alignment error of 2.4mm was reduced to 1.6mm after dynamic motion adjustments. Figure ~\ref{fig9a} highlights these improvements, showing significant reductions in positional discrepancy (from 4.0mm to 2.2mm), pipetting deviation (from 0.4mL to 0.2mL), and repeatability error (from ±2.8mm to ±1.2mm), demonstrating the enhanced precision and consistency of the upgraded system.
\subsection{Sample Preparation and Mixing}
GAMORA was evaluated for sample preparation tasks, including dispensing into multi-well plates and irregular containers. The system was required to operate within tight spatial constraints, ensuring liquid delivery without overflow or spillage. It achieved a placement accuracy of 0.3mm when dispensing into wells and maintained repeatability within 0.5mm across repeated transfers. For irregular container geometries, the initial deviation of 0.7mm was mitigated through software-based adjustments to the gripper’s positioning system, significantly improving precision during non-standard operations.
\subsection{Repetitive Operations and Workflow Effectiveness}
\begin{figure}[htbp]
\centerline{\includegraphics[width=\linewidth, keepaspectratio]{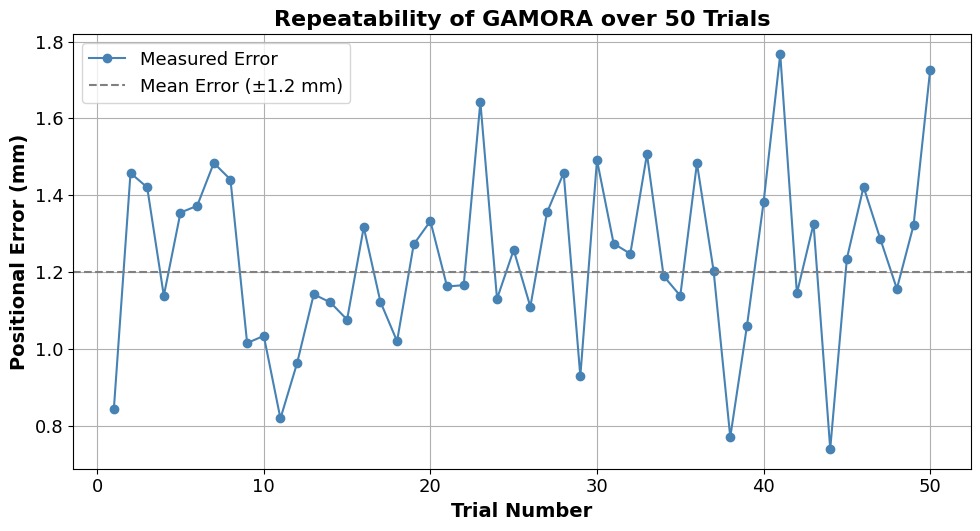}}
\caption{ Repeatability analysis of the GAMORA system over 50 trials}
\label{fig9}
\end{figure}

Repetitive sample handling and liquid transfer tasks are critical in virology labs. GAMORA was tested over 50 consecutive operation cycles to evaluate positional repeatability and performance stability. The system consistently executed high-throughput tasks with a positional error within ±1.2mm and an average cycle time of 45 seconds per 10-transfer set. The VR interface enabled precise control, ensuring low error accumulation over prolonged execution. Figure~\ref{fig9} illustrates the measured positional errors across 50 trials, showing fluctuations well within the defined mean error bounds and confirming the system's operational stability under repeated use.
\subsection{Planning and Execution Metrics}
GAMORA demonstrated improved motion planning efficiency, with planning time reduced from 0.75s to 0.5s and execution time shortened from 2.5s to 1.75s using MoveIt, while maintaining a 90–95\% path planning success rate. These gains ensured smoother operations with minimal mechanical jerk. The system also became more energy efficient. Idle current dropped from 250mA to 200mA, and peak current under heavy load halved from 2A to 1A. Correspondingly, power output was reduced from 100W to 50W, as shown in Table~\ref{fig:power_current_usage}. These gains significantly reduced thermal and electrical stress on components during extended operations.\begin{table}[htbp]
\centering
\caption{Current and power usage of original and improved system}
\label{fig:power_current_usage}
\begin{tabular}{|l|c|c|}
\hline
\textbf{Measurement Type} & \textbf{Original} & \textbf{Improved} \\
\hline
Idle Current (mA)      & 250 & 200 \\
\hline
Heavy Load Current (A) & 2   & 1   \\
\hline
Power Output (W)       & 100 & 50  \\
\hline
\end{tabular}
\end{table}
Resource usage was optimized, with CPU consumption dropping from 10\% to 5\% and RAM usage decreasing from 600MB to 550MB, enabling the system to execute complex tasks without straining computational resources. These hardware-level optimizations further enhance GAMORA’s suitability for long-duration, high-precision laboratory tasks.

\section{Conclusion}
The experimental results demonstrate that the GAMORA system achieves high precision and reliability across multiple laboratory automation tasks under controlled conditions. Quantitative evaluations showed that specimen handling accuracy improved significantly, with positional error reduced to 2.2mm and angular misalignment minimized to 2.5°. Pipetting operations yielded a volume deviation of only ±0.1mL from the 1mL target, with 95\% of transfers falling within the acceptable range. In multi-well plate preparation, the system maintained a placement accuracy of 0.3mm and demonstrated repeatability within 0.5mm across trials.Repetitive task execution over 50 cycles confirmed system stability, with positional deviation remaining within ±1.2mm. Planning and execution times were reduced to 0.5s and 1.75s, respectively, while achieving a path planning success rate of 90–95\%. The system also exhibited improvements in energy efficiency, reducing idle and heavy-load current consumption by 20–50\%, and halving overall power output. Computational resource usage remained low, with CPU usage reduced to 5\% and RAM to 550MB. GAMORA establishes a reliable, energy-efficient, and highly accurate framework for robotic automation of sensitive laboratory processes. Its ability to execute repeatable, fine-grained tasks with high success under real-world lab conditions marks a significant step toward minimizing human exposure in hazardous environments and optimizing laboratory throughput.

\bibliography{ref}

\begin{thebibliography}{10}
\providecommand{\url}[1]{#1}
\csname url@samestyle\endcsname
\providecommand{\newblock}{\relax}
\providecommand{\bibinfo}[2]{#2}
\providecommand{\BIBentrySTDinterwordspacing}{\spaceskip=0pt\relax}
\providecommand{\BIBentryALTinterwordstretchfactor}{4}
\providecommand{\BIBentryALTinterwordspacing}{\spaceskip=\fontdimen2\font plus
\BIBentryALTinterwordstretchfactor\fontdimen3\font minus \fontdimen4\font\relax}
\providecommand{\BIBforeignlanguage}[2]{{%
\expandafter\ifx\csname l@#1\endcsname\relax
\typeout{** WARNING: IEEEtran.bst: No hyphenation pattern has been}%
\typeout{** loaded for the language `#1'. Using the pattern for}%
\typeout{** the default language instead.}%
\else
\language=\csname l@#1\endcsname
\fi
#2}}
\providecommand{\BIBdecl}{\relax}
\BIBdecl

\bibitem{in1}
\BIBentryALTinterwordspacing
M.~A. Raza, M.~A. Ashraf, M.~N. Amjad, G.~U. Din, B.~Shen, and Y.~Hu, ``The peculiar characteristics and advancement in diagnostic methodologies of influenza a virus,'' \emph{Frontiers in Microbiology}, vol.~15, Jan. 2025. [Online]. Available: \url{http://dx.doi.org/10.3389/fmicb.2024.1435384}
\BIBentrySTDinterwordspacing

\bibitem{in2}
\BIBentryALTinterwordspacing
Harsh and P.~Tripathi, ``Medical viruses: diagnostic techniques,'' \emph{Virology Journal}, vol.~20, no.~1, Jul. 2023. [Online]. Available: \url{http://dx.doi.org/10.1186/s12985-023-02108-w}
\BIBentrySTDinterwordspacing

\bibitem{in3}
\BIBentryALTinterwordspacing
K.~Kim~Le and S.~D. Blacksell, ``Dengue virus biosafety: An analysis of evidence, global inconsistencies, and risk gaps,'' \emph{Applied Biosafety}, Mar. 2025. [Online]. Available: \url{http://dx.doi.org/10.1089/apb.2024.0065}
\BIBentrySTDinterwordspacing

\bibitem{in4}
\BIBentryALTinterwordspacing
E.~Zavaleta-Monestel, C.~Rojas-Chinchilla, A.~Anchía-Alfaro, D.~Quesada-Loría, J.~García-Montero, S.~Arguedas-Chacón, and G.~Hanley-Vargas, ``Tracking the threat, 50 years of laboratory-acquired infections: A systematic review,'' \emph{Acta Microbiologica Hellenica}, vol.~70, no.~2, p.~11, Mar. 2025. [Online]. Available: \url{http://dx.doi.org/10.3390/amh70020011}
\BIBentrySTDinterwordspacing

\bibitem{in5}
\BIBentryALTinterwordspacing
P.~Berche, ``Laboratory-associated infections and biosafety,'' \emph{La Presse Médicale}, p. 104277, Apr. 2025. [Online]. Available: \url{http://dx.doi.org/10.1016/j.lpm.2025.104277}
\BIBentrySTDinterwordspacing

\bibitem{in6}
\BIBentryALTinterwordspacing
S.~Ghose, ``Bolstering biosafety: Analyzing laboratory-acquired infections and biosafety protocols,'' 2024. [Online]. Available: \url{http://dx.doi.org/10.2139/ssrn.4864642}
\BIBentrySTDinterwordspacing

\bibitem{in7}
\BIBentryALTinterwordspacing
A.~Gao, R.~R. Murphy, W.~Chen, G.~Dagnino, P.~Fischer, M.~G. Gutierrez, D.~Kundrat, B.~J. Nelson, N.~Shamsudhin, H.~Su, J.~Xia, A.~Zemmar, D.~Zhang, C.~Wang, and G.-Z. Yang, ``Progress in robotics for combating infectious diseases,'' \emph{Science Robotics}, vol.~6, no.~52, Mar. 2021. [Online]. Available: \url{http://dx.doi.org/10.1126/scirobotics.abf1462}
\BIBentrySTDinterwordspacing

\bibitem{in8}
\BIBentryALTinterwordspacing
L.~K\"{o}niger, C.~Malkmus, D.~Mahdy, T.~D\"{a}ullary, S.~G\"{o}tz, T.~Schwarz, M.~Gensler, N.~Pallmann, D.~Cheufou, A.~Rosenwald, M.~M\"{o}llmann, D.~Groneberg, C.~Popp, F.~Groeber‐Becker, M.~Steinke, and J.~Hansmann, ``Rebia—robotic enabled biological automation: 3d epithelial tissue production,'' \emph{Advanced Science}, vol.~11, no.~45, Sep. 2024. [Online]. Available: \url{http://dx.doi.org/10.1002/advs.202406608}
\BIBentrySTDinterwordspacing

\bibitem{in9}
\BIBentryALTinterwordspacing
R.~Bogue, ``Robots in the laboratory: a review of applications,'' \emph{Industrial Robot: An International Journal}, vol.~39, no.~2, p. 113–119, Mar. 2012. [Online]. Available: \url{http://dx.doi.org/10.1108/01439911211203382}
\BIBentrySTDinterwordspacing

\bibitem{in10}
\BIBentryALTinterwordspacing
A.~Neri, M.~Coduri, V.~Penza, A.~Santangelo, A.~Oliveri, E.~Turco, M.~Pizzirani, E.~Trinceri, D.~Soriero, F.~Boero, S.~Ricci, and L.~S. Mattos, ``A novel affordable user interface for robotic surgery training: design, development and usability study,'' \emph{Frontiers in Digital Health}, vol.~6, Jul. 2024. [Online]. Available: \url{http://dx.doi.org/10.3389/fdgth.2024.1428534}
\BIBentrySTDinterwordspacing

\bibitem{in11}
\BIBentryALTinterwordspacing
L.~Turchet, F.~Gentilini, S.~Malandra, A.~Veccia, A.~Antonelli, and C.~Scoffone, ``Medical training in virtual reality: a gamification approach,'' \emph{Virtual Reality}, vol.~29, no.~2, May 2025. [Online]. Available: \url{http://dx.doi.org/10.1007/s10055-025-01159-4}
\BIBentrySTDinterwordspacing

\bibitem{in12}
\BIBentryALTinterwordspacing
P.~Strojny and N.~Dużmańska-Misiarczyk, ``Measuring the effectiveness of virtual training: A systematic review,'' \emph{Computers \& amp; Education: X Reality}, vol.~2, p. 100006, 2023. [Online]. Available: \url{http://dx.doi.org/10.1016/j.cexr.2022.100006}
\BIBentrySTDinterwordspacing

\bibitem{in13}
\BIBentryALTinterwordspacing
G.~Tao, B.~Garrett, T.~Taverner, E.~Cordingley, and C.~Sun, ``Immersive virtual reality health games:a narrative review of game design,'' \emph{Journal of NeuroEngineering and Rehabilitation}, vol.~18, no.~1, Feb. 2021. [Online]. Available: \url{http://dx.doi.org/10.1186/s12984-020-00801-3}
\BIBentrySTDinterwordspacing

\bibitem{in14}
V.~Angelov, E.~Petkov, G.~Shipkovenski, and T.~Kalushkov, ``Modern virtual reality headsets,'' in \emph{2020 International Congress on Human-Computer Interaction, Optimization and Robotic Applications (HORA)}, 2020, pp. 1--5.

\bibitem{in15}
\BIBentryALTinterwordspacing
M.~Mangalam, S.~Oruganti, G.~Buckingham, and C.~W. Borst, ``Enhancing hand-object interactions in virtual reality for precision manual tasks,'' \emph{Virtual Reality}, vol.~28, no.~4, Nov. 2024. [Online]. Available: \url{http://dx.doi.org/10.1007/s10055-024-01055-3}
\BIBentrySTDinterwordspacing

\bibitem{in16}
\BIBentryALTinterwordspacing
Z.~Ali, M.~F. Sheikh, A.~Al~Rashid, Z.~U. Arif, M.~Y. Khalid, R.~Umer, and M.~Ko\c{c}, ``Design and development of a low-cost 5-dof robotic arm for lightweight material handling and sorting applications: A case study for small manufacturing industries of pakistan,'' \emph{Results in Engineering}, vol.~19, p. 101315, Sep. 2023. [Online]. Available: \url{http://dx.doi.org/10.1016/j.rineng.2023.101315}
\BIBentrySTDinterwordspacing

\bibitem{in17}
E.~Rosen, D.~Whitney, E.~Phillips, D.~Ullman, and S.~Tellex, ``Testing robot teleoperation using a virtual reality interface with ros reality,'' 03 2018.

\bibitem{lr1}
\BIBentryALTinterwordspacing
P.~Schleer, P.~Kaiser, S.~Drobinsky, and K.~Radermacher, ``Augmentation of haptic feedback for teleoperated robotic surgery,'' \emph{International Journal of Computer Assisted Radiology and Surgery}, vol.~15, no.~3, p. 515–529, Jan. 2020. [Online]. Available: \url{http://dx.doi.org/10.1007/s11548-020-02118-x}
\BIBentrySTDinterwordspacing

\bibitem{lr2}
\BIBentryALTinterwordspacing
C.~Cruz~Ulloa, D.~Domínguez, J.~del Cerro, and A.~Barrientos, ``Analysis of mr–vr tele-operation methods for legged-manipulator robots,'' \emph{Virtual Reality}, vol.~28, no.~3, Jul. 2024. [Online]. Available: \url{http://dx.doi.org/10.1007/s10055-024-01021-z}
\BIBentrySTDinterwordspacing

\bibitem{lr18}
\BIBentryALTinterwordspacing
A.~Gao, R.~R. Murphy, W.~Chen, G.~Dagnino, P.~Fischer, M.~G. Gutierrez, D.~Kundrat, B.~J. Nelson, N.~Shamsudhin, H.~Su, J.~Xia, A.~Zemmar, D.~Zhang, C.~Wang, and G.-Z. Yang, ``Progress in robotics for combating infectious diseases,'' \emph{Science Robotics}, vol.~6, no.~52, Mar. 2021. [Online]. Available: \url{http://dx.doi.org/10.1126/scirobotics.abf1462}
\BIBentrySTDinterwordspacing

\bibitem{lr19}
S.~Kannan, K.~Subbaram, S.~Ali, and H.~Kannan, ``A systematic review with narrative synthesis on medical robotics and laboratory automation in the control of sars-cov-2, ebola and h1n1 (swine flu) viruses,'' \emph{Journal of Health and Social Sciences}, vol.~5, 05 2020.

\bibitem{lr20}
\BIBentryALTinterwordspacing
I.~Holland and J.~A. Davies, ``Automation in the life science research laboratory,'' \emph{Frontiers in Bioengineering and Biotechnology}, vol.~8, Nov. 2020. [Online]. Available: \url{http://dx.doi.org/10.3389/fbioe.2020.571777}
\BIBentrySTDinterwordspacing

\bibitem{lr3}
\BIBentryALTinterwordspacing
K.~Angers, K.~Darvish, N.~Yoshikawa, S.~Okhovatian, D.~Bannerman, I.~Yakavets, F.~Shkurti, A.~Aspuru-Guzik, and M.~Radisic, ``Roboculture: A robotics platform for automated biological experimentation,'' 2025. [Online]. Available: \url{https://arxiv.org/abs/2505.14941}
\BIBentrySTDinterwordspacing

\bibitem{lr4}
E.~Babaians, M.~Tamiz, Y.~Sarfi, A.~Mogoei, and E.~Mehrabi, ``Ros2unity3d; high-performance plugin to interface ros with unity3d engine,'' in \emph{2018 9th Conference on Artificial Intelligence and Robotics and 2nd Asia-Pacific International Symposium}, 2018, pp. 59--64.

\bibitem{lr11}
J.~Xie, ``Research on key technologies base unity3d game engine,'' in \emph{2012 7th International Conference on Computer Science \& Education (ICCSE)}, 2012, pp. 695--699.

\bibitem{lr12}
\BIBentryALTinterwordspacing
J.~Platt and K.~Ricks, ``Comparative analysis of ros-unity3d and ros-gazebo for mobile ground robot simulation,'' \emph{Journal of Intelligent \& amp; Robotic Systems}, vol. 106, no.~4, Dec. 2022. [Online]. Available: \url{http://dx.doi.org/10.1007/s10846-022-01766-2}
\BIBentrySTDinterwordspacing

\bibitem{lr13}
\BIBentryALTinterwordspacing
M.~Singh, J.~Kapukotuwa, E.~L.~S. Gouveia, E.~Fuenmayor, Y.~Qiao, N.~Murry, and D.~Devine, ``Unity and ros as a digital and communication layer for digital twin application: Case study of robotic arm in a smart manufacturing cell,'' \emph{Sensors}, vol.~24, no.~17, p. 5680, Aug. 2024. [Online]. Available: \url{http://dx.doi.org/10.3390/s24175680}
\BIBentrySTDinterwordspacing

\bibitem{lr5}
\BIBentryALTinterwordspacing
S.~K. Chinnasamy, H.~P. Sura, A.~Saleem, A.~Kathirvel, and P.~Rangan, ``Digital twin of robot manipulator using ros,'' in \emph{IV INTERNATIONAL SCIENTIFIC FORUM ON COMPUTER AND ENERGY SCIENCES (WFCES II 2022)}, vol. 2948.\hskip 1em plus 0.5em minus 0.4em\relax AIP Publishing, 2023, p. 040004. [Online]. Available: \url{http://dx.doi.org/10.1063/5.0178239}
\BIBentrySTDinterwordspacing

\bibitem{lr14}
\BIBentryALTinterwordspacing
E.~Dey, M.~Walczak, M.~S. Anwar, N.~Roy, J.~Freeman, T.~Gregory, N.~Suri, and C.~Busart, ``A novel ros2 qos policy-enabled synchronizing middleware for co-simulation of heterogeneous multi-robot systems,'' in \emph{2023 32nd International Conference on Computer Communications and Networks (ICCCN)}.\hskip 1em plus 0.5em minus 0.4em\relax IEEE, Jul. 2023, p. 1–10. [Online]. Available: \url{http://dx.doi.org/10.1109/ICCCN58024.2023.10230109}
\BIBentrySTDinterwordspacing

\bibitem{lr6}
\BIBentryALTinterwordspacing
X.~Wang, ``\BIBforeignlanguage{en}{Enabling human-robot partnerships in digitally-driven construction work through integration of building information models, interactive virtual reality, and process-level digital twins},'' 2022. [Online]. Available: \url{http://deepblue.lib.umich.edu/handle/2027.42/174292}
\BIBentrySTDinterwordspacing

\bibitem{lr15}
\BIBentryALTinterwordspacing
Y.~Lei, Z.~Su, and C.~Cheng, ``Virtual reality in human-robot interaction: Challenges and benefits,'' \emph{Electronic Research Archive}, vol.~31, no.~5, p. 2374–2408, 2023. [Online]. Available: \url{http://dx.doi.org/10.3934/era.2023121}
\BIBentrySTDinterwordspacing

\bibitem{lr16}
\BIBentryALTinterwordspacing
X.~Wang, L.~Shen, and L.-H. Lee, ``A systematic review of xr-enabled remote human-robot interaction systems,'' \emph{ACM Computing Surveys}, vol.~57, no.~11, p. 1–37, Jun. 2025. [Online]. Available: \url{http://dx.doi.org/10.1145/3730574}
\BIBentrySTDinterwordspacing

\bibitem{lr7}
\BIBentryALTinterwordspacing
E.~Rosen, D.~Whitney, E.~Phillips, G.~Chien, J.~Tompkin, G.~Konidaris, and S.~Tellex, ``Communicating and controlling robot arm motion intent through mixed-reality head-mounted displays,'' \emph{The International Journal of Robotics Research}, vol.~38, no. 12–13, p. 1513–1526, Apr. 2019. [Online]. Available: \url{http://dx.doi.org/10.1177/0278364919842925}
\BIBentrySTDinterwordspacing

\bibitem{lr8}
\BIBentryALTinterwordspacing
H.-P. Tsai, C.-W. Lin, Y.-J. Lin, C.-S. Yeh, and Y.-S. Shan, ``Novel software for high-level virological testing: Self-designed immersive virtual reality training approach,'' \emph{Journal of Medical Internet Research}, vol.~25, p. e44538, Jun. 2023. [Online]. Available: \url{http://dx.doi.org/10.2196/44538}
\BIBentrySTDinterwordspacing

\bibitem{lr9}
C.~Chan, A.~Pelosi, and A.~Brown, ``Vr controlled remote robotic teleoperation for construction applications,'' 2023.

\bibitem{lr10}
M.~C. Charão~dos Santos, V.~A. Sangalli, and M.~S. Pinho, ``Evaluating the use of virtual reality on professional robotics education,'' in \emph{2017 IEEE 41st Annual Computer Software and Applications Conference (COMPSAC)}, vol.~1, 2017, pp. 448--455.

\bibitem{lr17}
I.~Mehta, K.~Bimbraw, R.~G. Chittawadigi, and S.~K. Saha, ``A teach pendant to control virtual robots in roboanalyzer,'' in \emph{2016 International Conference on Robotics and Automation for Humanitarian Applications (RAHA)}, 2016, pp. 1--6.

\end{thebibliography}

\end{document}